\documentclass[11pt,a4paper]{article}
\usepackage[hyperref]{acl2021}
\usepackage{times}
\usepackage{latexsym}
\pdfoutput=1

\usepackage{microtype}

\aclfinalcopy % Uncomment this line for the final submission

\usepackage{times}

\usepackage{soul}
\usepackage{url}
\usepackage[utf8]{inputenc}
\usepackage{caption}
\usepackage{graphicx}
\usepackage{amsmath}
\usepackage{booktabs}
\usepackage{amsfonts}
\usepackage{soul}

\usepackage{xcolor}
\usepackage[linesnumbered,ruled,vlined]{algorithm2e}

\title{Zero-Shot Compositional Concept Learning}

\author{Guangyue Xu \\
  Michigan State University \\
  \texttt{xuguang3@msu.edu} \\\And
  Parisa Kordjamshidi \\
  Michigan State University \\
  \texttt{kordjams@msu.edu} \\\And
  Joyce Y. Chai \\
  University of Michigan\\
  \texttt{chaijy@umich.edu}\\
  }

\date{}

\begin{document}
\maketitle
\begin{abstract}
In this paper, we study the problem of recognizing compositional \textit{attribute-object} concepts within the zero-shot learning (ZSL) framework.
We propose an episode-based cross-attention (EpiCA) network which combines merits of cross-attention mechanism and episode-based training strategy to recognize novel compositional concepts.
Firstly, EpiCA bases on cross-attention to correlate \textit{concept-visual} information and utilizes the gated pooling layer to build contextualized  representations for both images and concepts.
The updated representations are used  for a more in-depth multi-modal relevance calculation for concept recognition.
Secondly, a two-phase episode training strategy, especially the transductive phase, is adopted to utilize unlabeled test examples to alleviate the low-resource learning problem.
Experiments on two widely-used zero-shot compositional learning (ZSCL) benchmarks have demonstrated the effectiveness of the model compared with recent approaches on both conventional and generalized ZSCL settings. 
\end{abstract}

\section{Introduction}

\begin{figure}
  \includegraphics[width=0.9\linewidth]{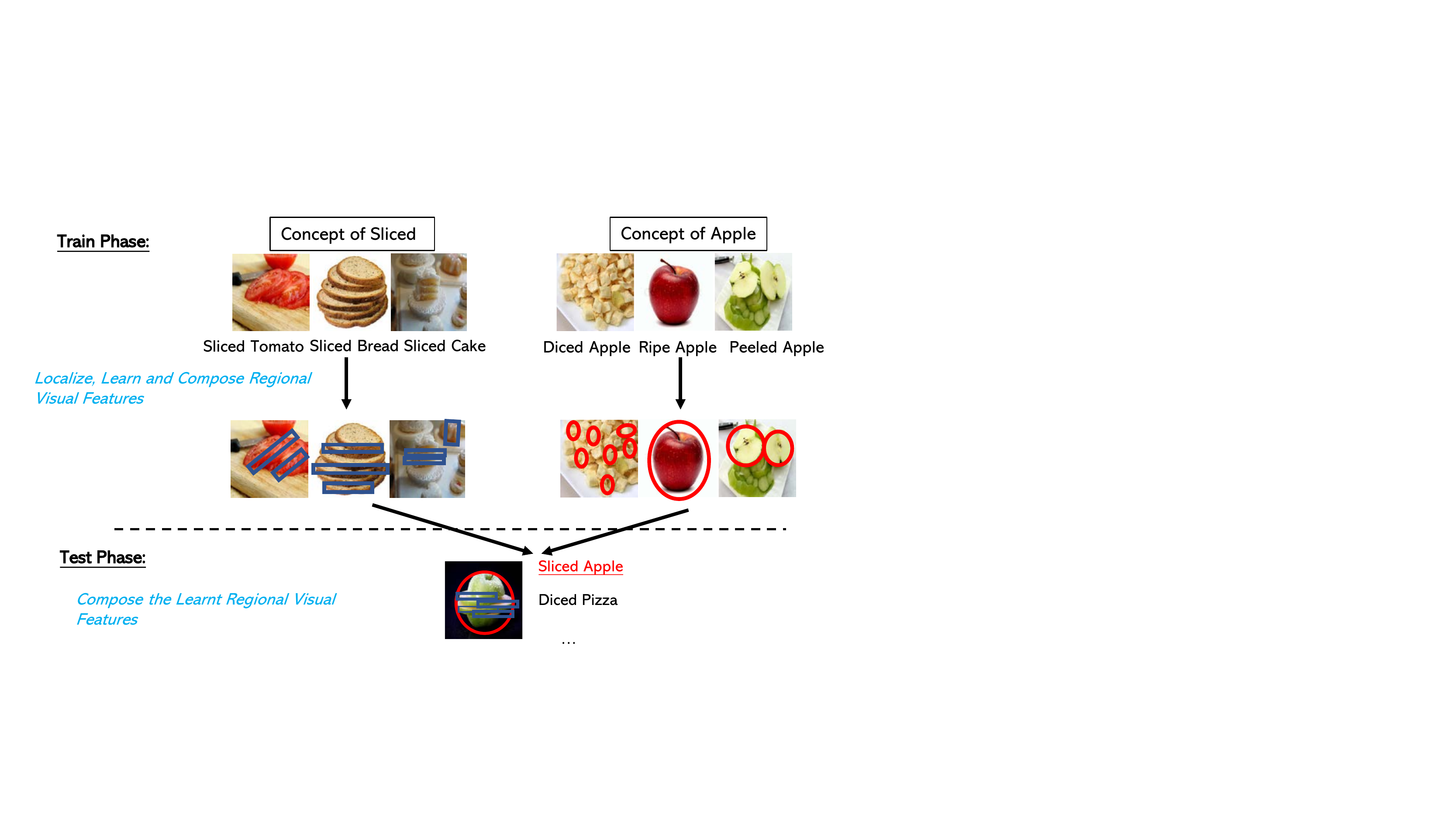}
  \caption{Given the concepts of \textit{sliced} and \textit{apple} in the training phase, our target is to recognize the novel compositional concept \textit{slice apple} which doesn't appear in the training set  by decomposing, grounding and composing concept-related visual features.}
  \label{fig:example}
\end{figure}

Humans can recognize novel concepts through composing previously learnt knowledge - known as compositional generalization ability~\cite{lake2015human,lake2018generalization}. As a key critical capacity to build modern AI systems, this paper investigates the problem of zero-shot  compositional learning (ZSCL) focusing on recognizing novel compositional \textit{attribute-object} pairs appeared in the images.  For example in Figure~\ref{fig:example}, suppose the training set has images with compositional concepts \textit{sliced-tomato, sliced-cake, ripe-apple, peeled-apple}, etc. Given a new image, our goal is to assign a novel compositonal concept \textit{sliced-apple} to the image by composing the element concepts, \textit{sliced} and \textit{apple}, learned from the training data. Although \textit{sliced} and \textit{apple} have appeared with other objects or attributes, the combination of this attribute-object pair is not observed in the training set.

This is a challenging problem, because objects with different attributes often have a significant diversity in their visual features. While \textit{red apple} has similar visual features as the \textit{apple} prototype, \textit{sliced apple} presents rather different visual features as shown in Fig~\ref{fig:example}. Similarly, same attributes can have different visual effects depending on the modified objects. For example, \textit{old} has different visual effect in objects of \textit{old town} compared to objects of \textit{old car}.

Despite recent progress~\cite{misra2017red,li2020symmetry}, previous works still suffer several limitations: %\jyc{not sure what you mean by lack of context. we know language BERT is context dependent, do you mean basic visual concept is a fixed vector without context??}
(1) Most existing methods adopt metric learning framework by projecting concepts and images into shared latent space, and focus on regularizing the structure of the latent space by adding principled constraints without considering the relationship between concepts and visual features.
Our work brings a new perspective, the relevance-based framework inspired by \citeauthor{sung2018learning}, to conduct compositional concept learning.
%(1) In previous works, concepts and images are represented by vectors without considering related information from the other modality and such general non-contextualized unimodal representation may not help learn the concept meaning.
(2)Previous works represent concept and image by the same vector regardless of the context it occurs.  However, cross concept-visual representation often provides more grounded information to help in recognizing objects and attributes which will consequently help in learning their compositions. %al concept learning task, which is understudied  in recent studies.
%\pk{ you mentioned suffer from limitations, but the three items you mention are not really limitations, specially this one here. and it seems all three are talking about the same thing, please think about it and revise.}

Motivated by the above discussions, we propose an Episode-based Cross Attention (EpiCA) network to capture multi-modal interactions and exploit the visual clues to learn novel compositional concepts.
Specifically, within each episode, we first adopt cross-attention encoder to fuse the concept-visual information and discover possible relationships between image regions and element concepts which corresponds to the localizing and learning phase in Fig.\ref{fig:example}.
%improve the representations for both modalities after the unimodal encoding.
Second, gated pooling layer is introduced to obtain the global representation by selectively aggregating the salient element features corresponding to Fig.~\ref{fig:example}'s composing phase.
Finally, relevance score is calculated based on the updated features to update EpiCA.
%so EpiCA can learn 'where' to attend when learning novel concepts.
%Based on the observation that concept is usually grounded with only one entity in the image as shown in Fig.~\ref{fig:example}, we extract fine-grained visual features using the last Resnet convolution layer instead of regional features extracted by Faster-RCNN~\cite{ren2015faster} as previous work~\cite{hou2019cross}, 

The contribution of this work can be summarized as follows:
1) Different from previous work, EpiCA has the ability to learn and ground the attributes and objects in the image by cross-attention mechanism. 
2) Episode-based training strategy is introduced to train the model. Moreover, we are among the first works to employ the transductive training to select confident unlabelled examples to gain knowledge about novel compositional concepts. 
3) Empirical results show that our framework achieves competitive results on two benchmarks in conventional ZSCL setting. In the more realistic generalized ZSCL setting, our framework significantly outperforms SOTA and achieves over $2\times$ improved performance on several metrics. 

%\jyc{($2\times$ part is not accurate. our framework significantly outperforms the STOA (in several settings achieves over $2\times$ of performance). } 

%\jyc{(I think you can add the range of performance gains to make it stronger: e.g. with 80\%-120\% (you need to calculate that) gains, and in several settings, doubling performance compared to SOTA}. 

\section{Related Work}

\noindent{\bf{Compositional Concept Learning.}}
As a specific zero-shot learning (ZSL) problem, zero-shot compositional learning (ZSCL) tries to learn complex concepts by composing element concepts. Previous solutions can mainly be categorized as: (1) classifier-based methods train classifiers for element concepts and combine the element classifiers to recognize compositional concepts~\cite{chen2014inferring,misra2017red,Li_2019_ICCV}. (2) metric-based methods learn a shared space by minimizing the distance between the projected visual features and concept features~\cite{nagarajan2018attrop,li2020symmetry}. (3) GAN-based methods learn to generate samples from the semantic information and transfer ZSCL into a traditional supervised classification problem~\cite{wei2019adversarial}.
%Different from previous frameworks, our work brings a novel relevance-based 

\noindent{\bf {Attention Mechanism.}}
The attention mechanism selectively use the salient elements of the data  to compose the data representation and is adopted in various visiolinguistic tasks. Cross attention is employed to locate important image regions for text-image matching~\cite{lee2018stacked}. Self-attention and cross-attention are combined at different levels to search images with text feedback~\cite{chen2020image}.
More recent works refer Transformer ~\cite{vaswani2017attention} to design various visiolinguistic attention mechanism~\cite{lu2019vilbert}. 
%\pk{I think you do not need to talk about attention in general but make sure to point to the paper on image attribute and Yanbei-chen et all}

\noindent{\bf {Episode-based Training.}}
The data sparsity in low-resource learning problems, including few-shot learning and zero-shot learning, makes the typical fine-tuning strategy in deep learning not adaptable, due to not having enough labeled data and the overfitting problem. Most successful approaches in this field rely on an  episode-based training scheme:  performing model optimization over batches of tasks instead of batches of data. Through training multiple episodes, the model is expected to progressively accumulate knowledge on predicting the mimetic unseen classes within each episode. Representative work includes Matching network~\cite{vinyals2016matching}, Prototypical network~\cite{snell2017prototypical} and RelNet~\cite{sung2018learning}. 

The related works to EpiCA are RelNet~\cite{sung2018learning} and cvcZSL~\cite{Li_2019_ICCV}. 
%RelNet and cvcZSL follow episode-based training to address ZSL problems. 
Compared with these methods, we have two improvements including an explicit way to construct episodes which is more consistent with the test scenario and a cross-attention module to fuse and ground more detailed information between the concept space and the visual space.
%In the proposed \textit{visual-concept} cross-attention framework, given an image and a sentence, it first attends to basic concepts with respect to each image region, and compares each image region to the attended information from from the basic concepts to decide the importance of the image regions (e.g., visual features grounded by each basic concept). Likewise, in the proposed \textit{concept-visual} formulation, it first attends to image with respect to each basic concept and then decides to pay more or less attention to each basic concept.
%Different from VAL which represents text using the last LSTM hidden vector, EpiCA represents concepts separately, attends to each basic concept with respect to image regions and weighted the representation by self-gated mechanism.
%Moreover, we are among the first works to employ transductive strategies to learn the compositional concepts. 

%\pk{if you can add how you are different from other cross-modal representations on the same data set that will be more clear about the difference of your work if you want to highlight this as a contribution}

\begin{figure*}[h]
\centering
  \includegraphics[width=1.0\linewidth,height=0.24\textheight]{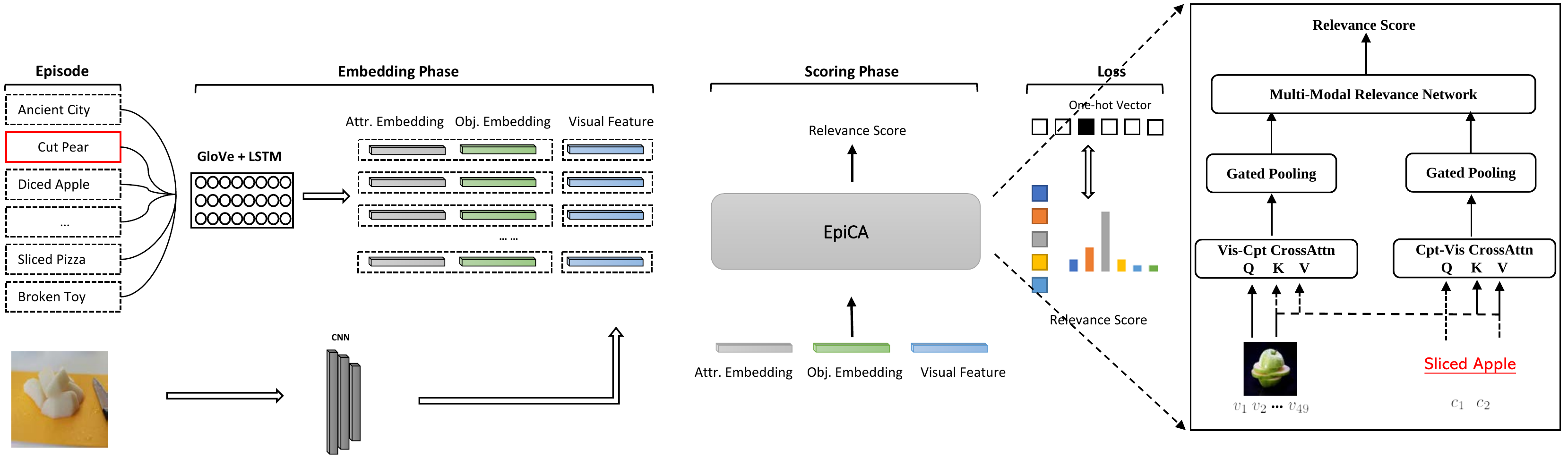}
  \caption{Illustration of the proposed EpiCA framework. It is a two-stage training framwork, including the inductive learning and the transductive learning. Both phases are trained on episodes illustrated in Alg.~\ref{alg:ins}.}
  \label{fig:arch}
\end{figure*}

\section{Approach}

\subsection{Task Definition}
Different from the traditional supervised setting where training concepts and test concepts are from the same domain, our problem focuses on recognizing novel compositional concepts of attributes and objects which are not seen during the training phase. Although we have seen all the attributes and objects in the training set, their compositions are novel \footnote{We refer concept as compositional concept, element concept as the attribute and the object in the rest of the paper.}.

We model this problem within the ZSL framework where the dataset is divided into the seen domain $\mathcal{S}=\{({v}_{s},{y}_{s}) | {v}_{s}\in\mathcal{V}^{s}, {y}_{s} \in\mathcal{Y}^{s}\}$ for training and the unseen domain $\mathcal{U}=\{({v}_{u},{y}_{u}) | {v}_{u} \in \mathcal{V}^{u}, {y}_{u} \in\mathcal{Y}^{u}\}$ for test, where ${v}$ is the visual feature of image $\mathcal{I}$ which can be extracted using deep convolution networks and $y$ is the corresponding label which consists of an  attribute label ${a}$ and a object label ${o}$ as $y=(a,o)$ satisfying ${a}_{u} \subseteq {a}_{s}$, ${o}_{u} \subseteq {o}_{s}$ and $\mathcal{Y}_{s} \cap \mathcal{Y}_{u}=\phi$. 
Moreover, we address the problem in both conventional ZSCL setting and generalized ZSCL setting.
In conventional ZSCL, we only consider unseen pairs in the test phase and the target is to learn a mapping function $\mathcal{V} \mapsto \mathcal{Y}^{u}$.
In generalized ZSCL, images with both seen and unseen concepts can appear in the test set, and the mapping function changes to
$\mathcal{V} \mapsto \mathcal{Y}^{s} \cup \mathcal{Y}^{u}$ which is a more general and realistic setting.

\subsection{Overall Framework}
As summarized in Fig.~\ref{fig:arch}, EpiCA consists of the cross-attention encoder, gated pooling layer and multi-modal relevance network to compute the relevance score between concepts and images. In order to accumulate the knowledge between images and  concepts, EpiCA is trained by episodes including the following two phases:

\begin{itemize}
    \item \textit{Inductive training phase} constructs episodes from the seen concepts and trains EpiCA based on these constructed episodes.
    \item \textit{Transductive training phase}  employs the self-taught methodology to collect confident pseudo-labeled test items to further fine-tune EpiCA.
\end{itemize}

\subsection{Unimodal Representation}

\textbf{Concept Representation.}
Given a compositonal concept $(a,o)$, we first transform attribute and object using $300$-D \textit{GloVe}~\cite{pennington2014glove}  separately. Then we use one layer BiLSTM~\cite{hochreiter1997long} to obtain contextualized representation for concepts with $d_k$ hidden units. Instead of using the final state, we maintain the output features for both attribute and object and output feature matrix $C \in \mathbb{R}^{2 \times d_k}$ for each compoisitonal concept.

\noindent\textbf{Image Representation.}
We extract the visual features using pretrained ResNet~\cite{he2016deep} from a given image. In order to obtain more detailed visual features for concept recognition, we keep the output from the last convolutional layer of ResNet-$18$ to represent the image and therefore each image is split into $7\times7=49$ visual blocks with each block as a $512$-dim vector denoted as $\mathbf{V}=\left(\mathbf{v}_{1}, \mathbf{v}_{2}, \ldots, \mathbf{v}_{49}\right)$. Each element represents a region in the image. We further convert $v_i$ with a linear transformation $v_i=\mathbf{W}^{\top} v_i$, where $\mathbf{W} \in \mathbb{R}^{512 \times d_k}$ is the weight matrix to transfer the image into the joint concept-image space.

\subsection{Cross Attention Encoder}
\textbf{Motivation.} Previous works usually utilize vector representation for both concepts and images and construct a metric space by pushing aligned images and concepts closer to each other. The potential limitation of such frameworks is that the same vector representations without context information will miss sufficient detailed information needed for grounding and recognizing objects and attributes appeared in the images. 
We observe that certain visual blocks in the image can be more related to certain element concept and certain element concept may highlight different visual blocks. Inspired by this observation, our model addresses the previous limitation by introducing cross-attention encoder and constructs more meaningful cross-modality representation for both images and element concepts for compositional concept recognition.

\textbf{Cross Attention Layer.}
To fuse and ground information between visual space and concept space, we first design a correlation layer to calculate the correlation map between the two spaces, which is used to guide the generation of the cross attention map. 
Given an image and a candidate concept, after extracting unimodal representations, the  correlation layer computes the semantic relevance between visual blocks $\left\{v_{i}\right\}_{i=1}^{49}$ and element concepts  $\left\{c_{j}\right\}_{j=1}^{2}$ \footnote{Each compositional concept only has two elements, attribute and object.} with cosine distance   and output the final \textit{image-to-concept} relevance matrix as $R \in \mathbb{R}^{49 \times 2}$ with each element $r_{ij}$ calculated using Eq.~\ref{eq:cos_attn}. We can easily have another \textit{concept-to-image} relevance matrix by transposing $R$. 
%We take image input as query $Q$ and concept input as both key $K$ and value $V$  where $K=V$  in the following illustration.  

\begin{equation}
r_{i j}=\left(\frac{v_{i}}{\left\|v_{i}\right\|_{2}}\right)^{T}\left(\frac{c_{j}}{\left\|c_{j}\right\|_{2}}\right), i \in[1, 49], j \in[1, 2]
\label{eq:cos_attn}
\end{equation}

In order to obtain attention weights, we need to normalize the relevance score $r_{ij}$ as Eq.~\ref{eq:attn} as \cite{relu_attn}.

\begin{equation}
\bar{r}_{i j}=\frac{\operatorname{relu}\left(r_{i j}\right)}{\sqrt{\sum_{j=1}^n \operatorname{relu}\left(r_{i j}\right)^{2}}}
\label{eq:attn}
\end{equation}

%We have query space $Q$ and  context space where $K=V$ in our setting 

After obtaining the normalized attention score, we can calculate the cross-attention representation 
based on the selected query space $Q$ and the context space $V$, where $V=K$ in our setting as shown in Fig.~\ref{fig:arch}. Taking \textit{image-to-concept} attention for example, given a visual block feature $v_i$ as query, cross attention encoding is performed over the element concept space $C$ using Eq.~\ref{eq:final_attn}.

\begin{equation}
\widehat{v}_{i}=\sum_{j=1}^{n} \alpha_{i j} c_{j}, \quad \text{ s.t. } \quad \alpha_{i j}=\frac{\exp \left(\lambda \bar{r}_{i j}\right)}{\sum_{j=1}^{n} \exp \left(\lambda \bar{r}_{i j}\right)}
\label{eq:final_attn}
\end{equation}

\noindent where $\lambda$ is the inverse temperature parameter of the softmax function~\cite{chorowski2015attention} to control the smoothness of the attention distribution.

\textbf{Visually-Attended Concept Representation.}
The goal of this module is to align and represent concepts with related visual blocks and help further determine the alignment between element concepts and image regions. We use concept embedding as query and collect visual clues using Eq.~\ref{eq:final_attn} and the final visually-attended features for compositional concept is $\widehat{c} \in R^{2 \times d_{k}}$.

%\begin{equation}
%\begin{array}{l}
%e_{l}=\text { embedding }\left(w_{l}\right) \\
%\vec{h}_{l}=\overrightarrow{LSTM}\left(e_{l}, %\vec{h}_{l-1}\right) \\
%\overleftarrow{h}_{l}=\overleftarrow{L S T M}\left(e_{l}, \overleftarrow{h}_{l+1}\right) \\
%h_{l}=\left[\vec{h}_{l}, \overleftarrow{h}_{l}\right]
%\end{array}
%\end{equation}

%where $\rightarrow$ and $\leftarrow$ denote the forward and backward directionsin bidirectional LSTM, espectively. $h_{l} \in \mathbb{R}^{C_{r}}$ is used as the vector representation for the base concept, which concatenates the hidden states of bidirectional LSTM at every time step (word) $l$ with  consideration  of  both  previous  and  future  relations  of words.

%\begin{equation}
%\begin{array}{l}
%\widehat{c}=\operatorname{Cross\_Attn}\left(c, v\right)
%\end{array}
%%\label{eq:final_cpt}
%\end{equation}
%\noindent where  representing the visual-attended attribute features and the attended object features with the same latent space dimension $d_h$ separately. 

\textbf{Concept-Attended Visual Representation.} 
An image representation grounded with element concept would be beneficial for compositional concept learning. Following the similar procedure as visually-attended concept representation, we take visual block features as query and concept embedding as context. We can calculate the concept-attended visual representation using Eq.~\ref{eq:final_attn}. The final result $\widehat{v} \in \mathbb{R}^{49 \times d_{k}}$ represents the concept-attended block visual features with the latent space dimension $d_k$. 
%\begin{equation}
%\begin{array}{l}
%\widehat{v}=\operatorname{Cross\_Attn}\left(v, c\right)\\
%\end{array}
%\label{eq:final_vis}
%\end{equation}

%\begin{figure}
%    \centering
%    \includegraphics[width=0.4\textwidth]{img/gate_sum.pdf}
%    \caption{For each input $X$ with elements $X_1, X_2, \ldots, X_N$}, gated pooling is used to weighted-sum the elements.
%    \label{fig:gate}
%\end{figure}

\subsection{Gated Pooling Layer}

After the cross-attention encoder, the output image features $V=\left[v_{1}, \ldots, v_{49}\right] \in \mathbb{R}^{49 \times d_k}$ and concept features $C=\left[c_{1}, c_{2}\right] \in \mathbb{R}^{2 \times d_k}$ are expected to contain  rich cross-modal information. Our target of gated pooling layer is to combine elements to form the final representation for concepts and images separately.
Pooling techniques can be directly deployed to obtain such representation. However, we argue that elements should have different effect on the final concept recognition. For example, background visual blocks shouldn't be paid much attention during concept recognition.
To address the assumption, we propose gated pooling layer to learn the relative importance of each element and dynamically control the contribution of each element in the final representation.
Specially, We apply one linear layers with parameter $W \in \mathbb{R}^{d_k \times 1}$ on the element feature $x_i$ and normalize  the  output  to  calculate  an  attention  weight $\alpha_i$ that indicates the relative importance of each element using Eq.~\ref{combine}.

\begin{equation}
\begin{array}{l}
x=\sum_{i}\alpha_{i} x_{i} \quad \text { s.t. } \quad\alpha_{i}=\frac{\exp \left(\left(W x_{i}\right)\right)}{\sum_{k=1}^{N} \exp \left(\left(W x_{k}\right)\right)}\\
\end{array}
\label{combine}
\end{equation}

\subsection{Multi-Modal Relevance Network}
After obtaining the updated features for both images $\widehat{v}_i$ and concepts $(\widehat{a},\widehat{o})_j$, we introduce the multimodal relevance network shared the spirit as \cite{sung2018learning} to calculate the relevance score as shown in Eq.~\ref{eq:score_func} 

\begin{equation}
s_{i, j}=g_{\phi}\left(\operatorname{concat}[(\widehat{v}_i), (\widehat{a},\widehat{o})_j]\right)
\label{eq:score_func}
\end{equation}

\noindent where $g$ is the relevance function implemented by two layer feed-forward network with trainable parameters $\phi$.

In order to train EpiCA, we add Softmax activation on the relevance score to measure the probability of image $i$ belonging to concept $j$ within the current episode as Eq.~\ref{eq:score_func} and update EpiCA  using cross-entropy loss.

\begin{equation}
p_j(\widehat{v}_i)=\frac{\exp (s_{i,j})}{\sum_{k=1}^{C} \exp \left(s_{i,k}\right)}
\label{eq:ce}
\end{equation}

%\begin{equation}
%\begin{aligned}
%p_j(v_i)&=\frac{\exp \left(r_{i,j} \left(\widehat{v}_i, (\widehat{a},\widehat{o})_{j} \right)\right)}{\sum_{k=1}^{C} \exp \left(r_{i,k} \left(\widehat{v}_i, (\widehat{a},\widehat{o})_{k} \right)\right)}\\
%L_s&=\sum_{j=1}^{C} \mathbbm{1}_{(a,o)_{i, j}} \log \left(p_{j}\left(v_i\right)\right)
%\end{aligned}
%\label{eq:seen_ce}
%\end{equation}

%\noindent where $C$ is the number of pairs in the current episode and $\mathbbm{1}_{(a,o)_{i, j}} \in \{0,1\}$ indicates whether image $i$ is matched with pair $j$ within current episode.

\begin{algorithm}[t]
\DontPrintSemicolon
    \SetKwFunction{isOddNumber}{isOddNumber}
    \KwIn{$\mathcal{D}_{train}=\{(v_m, (a_m,o_m)\}_{m=1}^{|Tr|}$, 
    $\mathcal{D}_{test}=\{v_n\}_{i=n}^{|Ts|}$, 
    task size $S$,
    sample interval $t$
    }
    \KwOut{Multi-Modal Rel. Function $f$}
    \tcp{Inductive Learning Phase}
    \For{$epoch \leftarrow 1$ \KwTo $E_{ind\_max}$}
    {
        \For{each image and the corresponding pair in the training set}
        {
            \textbf{Construct} an episode $[v_p, (a_p,o_p), (a_{n_1},o_{n_1}), \cdots, (a_{n_s},o_{n_s})]$.\\
            \textbf{Gated Cross-Attention Encoding} using Eq.~\ref{eq:cos_attn}, \ref{eq:attn}, \ref{eq:final_attn} and  
            \ref{combine}\\
            \textbf{Calculating} multi-modal relevance score using Eq~\ref{eq:score_func}. \\
            \textbf{Updating} EpiCA.\\
        }
    }
    \tcp{Transductive Learning Phase}
    \For{$epoch \leftarrow 1$ \KwTo $E_{trans\_max}$}
    {
        \If{$epoch\ \% \ t == 0$}
        {
           \textbf{Pick} confident samples from unseen set by Eq.~\ref{eq:thresh}.
        }
       \textbf{Updating} EpiCA  by Eq~\ref{eq:trans_loss}.
    }
\caption{Training EpiCA for ZSCL}
\label{alg:ins}
\end{algorithm}

\subsection{Training and Prediction}

\textbf{Inductive Training\label{sec:ind}.}
For each image and the corresponding pair label, we randomly sample negative pairs to form an episode which consists of an image $v_p$, a positive pair $(a_p,o_p)$ and a predefined number $n_t$ of negative pairs in the form of $[v_p, (a_p,o_p), (a_{n_1},o_{n_1}), \cdots, (a_{n_t},o_{n_t})]$. Then within each episode, we calculate the relevance score between  image and all candidate pairs using Eq.~\ref{eq:score_func}. Finally, we calculate the cross entropy loss using Eq.~\ref{eq:ce} and update EpiCA as shown in Alg.~\ref{alg:ins}.

\textbf{Transductive Training.}
The disjointness of the seen/unseen concept space will result in domain shift problems and cause the  predictions biasing towards seen concepts as pointed by~\cite{pan2009survey}.
Transductive training utilizes the unlabeled test set to alleviate the problem~\cite{dhillon2019baseline}. Specifically, transductive training has a sampling phase to select confident test samples and utilize the generalized cross entropy loss as Eq.~\ref{eq:gen_entropy} to update EpiCA.
%To address this problem, we follow the transductive training framework and employ the self-taught strategy to explore the information hidden in the test dataset.

Following previous work~\cite{li2019learning}, we use threshold-based method as Eq.~\ref{eq:thresh} to pick up confident examples.

\begin{equation}
\frac{p_{1}(\widehat{v}_i)}{p_{2}(\widehat{v}_i)} > \gamma
\label{eq:thresh}
\end{equation}

\noindent where $p$ is calculated by Eq.~\ref{eq:ce} and the threshold is the fraction of the highest label probability $p_{1}(\widehat{v}_i)$ and the second highest label probability $p_{2}(\widehat{v}_i)$ which measures the prediction peakiness in current episode. Only confident instances are employed to update EpiCA which is controlled by $\gamma$.

Moreover, the recently proposed generalized cross-entropy loss~\cite{zhang2018generalized} is used to calculate the loss for pseudo-labeled test examples as Eq.~\ref{eq:gen_entropy}. 

\begin{equation}
\mathcal{L}_{u}=\sum_{\left(v_i, (a,o)_{j}\right) \in \mathcal{U}} \frac{1-(p_j(\widehat{v}_i))^{q}}{q}
\label{eq:gen_entropy}
\end{equation}

\noindent where $p_j(\widehat{v}_i)$ is the probability of $\widehat{v}_i$ belonging to pair $(\widehat{a},\widehat{o})_j$ calculated using Eq.~\ref{eq:ce}. $q \in(0,1]$ is the hyper-parameter related to the noise level of the pseudo labels, with higher noisy pseudo labels requiring larger $q$.

Finally, the transductive loss is calculated as Eq.~\ref{eq:trans_loss}, where $\mathcal{L}_{u}$ corresponds to the generalized cross entropy loss from pseudo-labeled test examples and  $\mathcal{L}_{s}$ is the cross entropy loss for the training examples
\begin{equation}
\mathcal{L} = \mathcal{L}_{u} + \mathcal{L}_{s}.
\label{eq:trans_loss}
\end{equation}

\noindent\textbf{Prediction.} Given a new image with extracted feature $v_i$, we iterate over all the candidate pairs and select the pair with the highest relevance score as  $(\hat{a}, \hat{o})=\operatorname{argmax}_{\hat{a}, \hat{o}} s_{i,j}(\hat{v}_i , (\hat{a},\hat{o})_j)$ as Eq.~\ref{eq:score_func} using EpiCA.

\section{Experiments\label{sec:exp}}

\noindent\textbf{Dataset.}
We use similar dataset as in ~\cite{nagarajan2018attrop,purushwalkam2019taskdriven} for both conventional and generalized ZSCL settings with the split shown in Tab.~\ref{table:data_split}. Notably, generalized ZSCL setting has additional validation set for both benchmarks which allows cross-validation to set the hyperparameters. The generalized ZSCL evaluates the models on both seen/unseen sets.
\begin{itemize}
    \item{MIT-States}~\cite{isola2015discovering} has $245$ objects and $115$ attributes. In conventional ZSCL, the pairs are split into two disjoint sets with $1200$ seen pairs and $700$  unseen pairs. In generalized ZSCL, the validation set has $600$ pairs with $300$ pairs seen in the training set and $300$ pairs unseen during training and the test set has $800$ pairs with $400$ pairs seen and remaining $400$ pairs unseen in the training set. 
    \item{UT-Zappos}~\cite{yu2017semantic} contains images of $12$ shoe types as object labels and  $16$ material types as attribute labels. In conventional ZSCL, the dataset is split into disjoint seen set with $83$ pairs and unseen set with $33$ pairs. In generalized ZSCL, the $36$ pairs in the test set consists $18$ seen and $18$ unseen pairs. $15$ seen pairs and $15$ unseen pairs composes the validation set.  
\end{itemize}

\noindent\textbf{Implementation Details.}
We develop our model based on \textit{PyTorch}.
For all experiments, we adopt ResNet-$18$ pre-trained on ImageNet as the backbone to extract visual features. For \textit{attr-obj} pairs, we encode attributes and objects with $300$-dim  \textit{GloVe} and fix it during the training process. We randomly sample $50$ negative pairs to construct episodes. We use Adam with $10^{-3}$ as the initial learning rate and multiply the learning rate by $0.5$ every $5$ epoch and train the network for total $25$ epochs. We report the accuracy at the last epoch for conventional ZSCL. For generalized ZSCL, the accuracy is reported based on the validation set. Moreover, the batch size is set to $64$, $\lambda$ in Eq.~\ref{eq:final_attn} is set to $9$, $q$ in Eq.~\ref{eq:gen_entropy}  is set to $0.5$ and the threshold in Eq.~\ref{eq:thresh} is set to $10$. 
%and the implementation in ~\cite{nagarajan2018attrop} \footnote{https://github.com/Tushar-N/attributes-as-operators}.

\noindent\textbf{Baselines.} 
We compare EpiCA with the following SOTA methods:
1) Analog~\cite{chen2014inferring} trains a linear SVM classifier for the seen pairs and utilizes Bayesian Probabilistic Tensor Factorization to infer the unseen classifier weights. 
2) Redwine~\cite{misra2017red} leverages the compatibility between visual features $v$ and concepts semantic representation to do the recognition. 
3) AttOperator~\cite{nagarajan2018attrop} models composition by treating attributes as matrix operators to modify object state to score the compatibility.
4) GenModel~\cite{nan2019recognizing} adds reconstruction loss to boost the metric-learning performance.
% 5) TMN~\cite{purushwalkam2019taskdriven} generalize a set of base fully-connected(FC) modules to unseen pairs via re-weighting theseFC modules.
5) TAFE-Net~\cite{wang2019tafe} extracts visual features based on the pair semantic representation and utilizes a shared classifier to recognize novel concepts. 
6) SymNet~\cite{li2020symmetry} builds a transformation framework and adds group theory constraints to its latent space to recognize novel concepts.
We report the results according to the papers and the released official code
\footnote{https://github.com/Tushar-N/attributes-as-operators} 
\footnote{https://github.com/ucbdrive/tafe-net.git}
of the aforementioned baselines. 

\begin{table}[]
\centering
\small
\setlength\tabcolsep{3.5pt}
\begin{tabular}{*{5}c}
\toprule
&  \multicolumn{2}{c}{Conventional ZSCL} & \multicolumn{2}{c}{Generalized ZSCL} \\
&  MIT-States & Zappos & MIT-States & Zappos\\
\hline
\# Attr. & $115$ & $16$ & $115$ & $16$ \\
\# Obj. & $245$ & $12$ & $245$  & $12$\\
\hline
\# Train Pair & $1262$ & $83$ & $1262$ & 83\\
\# Train Img. & $34562$ & $24898$ & $30338$ & 22998\\
\hline
\# Test Pair & $700$  & $33$ & $800$ & 36\\
\# Test Img. & $19191$ & $4228$ & $12995$ & 2914\\
\hline
\# Val. Pair &  & & $600$ &  30\\
\# Val. Img. & & & $10420$ & 3214\\
\hline
\end{tabular}
\caption{Conventional and Generalized Data Split for MIT-States and Zappos Datasets.}
\label{table:data_split}
\end{table}

\begin{table}[ht]
\centering
\small
\setlength\tabcolsep{2pt}
\begin{tabular}{ccc}
\hline
Methods &  MIT-States(\%) & UT-Zappos(\%)\\ 
\hline \hline
Random & 0.14 & 3.0 \\
ANALOG  & 1.4 & 18.3 \\
%SAE & 14.0 & 31.0 \\
REDWINE & 12.5 & 40.3 \\
ATTOPERATOR & 14.2 & 46.2 \\
GenModel & 17.8 & 48.3 \\
TAFE-Net & 16.4 & 33.2 \\
SymNet & \textbf{19.9} & 52.1 \\
%\hline
%ECL(Inductive) & 14.13 & 48.99 \\
%ECL(Transductive) & 16.01 & 52.17 \\
\hline
EpiCA(Inductive) & 15.68 & \textbf{52.56} \\
EpiCA(Transductive) & 18.13 & \textbf{55.48} \\
\hline
\end{tabular}
\caption{Results of Conventional ZSCL setting}
\label{tab:close_result}
\end{table}

\begin{table*}[!htbp]
\centering
\small
\setlength\tabcolsep{5pt}
\begin{tabular}{*{13}c}
\toprule
&  \multicolumn{6}{c}{Mit-States} & \multicolumn{6}{c}{UT-Zappos} \\
&  \multicolumn{3}{c}{Val AUC} & \multicolumn{3}{c}{Test AUC} & \multicolumn{3}{c}{Val AUC} & \multicolumn{3}{c}{Test AUC}\\
Model Top k $\xrightarrow{}$ & 1 & 2 & 3 & 1 & 2 & 3 & 1 & 2 & 3 & 1 & 2 & 3\\
\midrule
AttOperator & 2.5 & 6.2 & 10.1 & 1.6 & 4.7 & 7.6 & 21.5 & 44.2 & 61.6 & 25.9 & 51.3 & 67.6 \\
RedWine &   2.9 & 7.3 & 11.8 & 2.4 & 5.7 & 9.3 & 30.4 & 52.2 & 63.5 & 27.1 & 54.6 & 68.8 \\
LabelEmbed+ & 3.0 & 7.6 & 12.2 & 2.0 & 5.6 & 9.4 & 26.4 & 49.0 & 66.1 & 25.7 & 52.1 & 67.8 \\
TMN & 3.5 & 8.1 & 12.4 & 2.9 & 7.1 & 11.5 & 36.8 & 57.1 & 69.2 & 29.3 & \textbf{55.3} & 69.8\\
SymNet & 4.3 & 9.8 & 14.8 & 3.0 & 7.6 & 12.3 & \verb|\| & \verb|\| & \verb|\| & \verb|\| & \verb|\| & \verb|\| \\
\midrule
%\textbf{ECL(Previous)} & \textbf{4.45}&\textbf{10.47}&\textbf{15.56}&\textbf{3.41}&\textbf{8.57}&\textbf{13.5}&\textbf{27.92}&\textbf{49.58}&\textbf{65.61}&\textbf{29.32}&\textbf{54.27}&\textbf{67.86}\\
%\textbf{Inductive ECL} & 6.71 & 12.03 & 21.51 &  6.12 & 12.23 & 19.87 & 25.43 & 49.18 & 61.50 & 25.63 & 46.73 & 60.15\\
%\textbf{Transductive ECL} & \textbf{8.54}&15.73&\textbf{22.89}&\textbf{6.45}&\textbf{12.84}&\textbf{20.98}&\textbf{52.47}&\textbf{67.13}&\textbf{78.25}&\textbf{34.14}&52.09&67.09\\
%\hline
\textbf{Inductive EpiCA} & 7.73 & 12.19 & 22.93 & 6.55 & 13.07 & 20.01 & 25.13 & 50.19 & 61.97 & 25.59 & 50.06 & 63.08\\
\textbf{Transductive EpiCA} &\textbf{9.01} & \textbf{17.63} & \textbf{24.01} & \textbf{7.18} & \textbf{14.02} & \textbf{21.31} & \textbf{53.18} & \textbf{68.71} & \textbf{77.89} & \textbf{35.04} & 54.83 & \textbf{70.02}\\
\bottomrule
\end{tabular}
\caption{AUC in percentage (multiplied by 100) on MIT-States and UT-Zappos. Our \textit{EpiCA} model outperforms the previous
methods by a large margin on MIT-States based on most of the metrics on UT-Zappos.}
\label{tab:auc_result}
\end{table*}

\subsection{Conventional ZSCL Setting}

\noindent\textbf{Quantitive Results.}
Top-1 accuracy metric is reported in this setting to compare different methods.
The top-1 accuracy of the unseen \textit{attr-obj} pairs for conventional ZSCL is presented in Tab.~\ref{tab:close_result}. EpiCA outperforms all baselines on Zappos benchmark and exceeds the state-of-the-art by $3.3\%$. It achieves comparable performance on MITStates benchmark. We will empirically analyze the model's behavior in later sections.

\subsection{Generalized ZSCL Setting}

In this setting, following the related work~\cite{purushwalkam2019taskdriven}, we measure the  performance with AUC metric. AUC introduces the concept of calibration bias which is a scalar value added to the predicting scores of unseen pairs. By changing the values of the calibration bias, we can draw an accuracy curve for seen/unseen sets. The area below the curve is the AUC metric as a measurement for the generalized ZSCL system.

\noindent\textbf{Quantitative results.} Tab.~\ref{tab:auc_result} provides comparisons between our EpiCA model and the previous methods on both the validation and testing sets. As Tab.~\ref{tab:auc_result} shows, the EpiCA model outperforms the previous methods by a large margin. 
On the challenging MIT-States dataset which has about $2000$ attribute-object pairs, all the baseline methods have a relatively low AUC score while our model is able to double the performance of the previous methods, indicating its effectiveness.

\subsection{Ablation Study}

We conduct ablation study on EpiCA and compare its performance in different settings.

\noindent\textbf{Importance of Transductive Learning.}
%Previous ZSCL works utilize regularization techniques or external knowledge to address the domain-shift problem. More direct solution, transductive learning, is employed in this work. 
The experimental results in Tab.~\ref{tab:close_result} and Tab.~\ref{tab:auc_result} show the importance of transductive learning. %Transductive learning improves the performance in both conventional ZSCL and generalized ZSCL. 
There are about $2\%$ and $3\%$ performance gains for MIT-States and UT-Zappos in conventional ZSCL. A significant improvement is observed for both datasets in generalized ZSCL. 
This is within our expectation because 1) our inductive model has accumulated knowledge about the elements of the concept and has the ability to pick confident test examples.  2) after training the model with the confident pseudo-labeled test data, it acquires the knowledge about unseen concepts. 
%Therefore an improved performance of ZSCL is more likely to be achieved.

\noindent\textbf{Importance of Cross-Attention (CA) Encoder.}
To analyze the effect of CA encoder, we remove CA (w/o CA) and use unimodal representations for both concepts and images. From Tab.~\ref{tab:abla}, it can be seen that EpiCA does depend on multi-modal information to do concept recognition and the results also verifies the rationale to fuse multi-modal information by cross-attention mechanism.

\noindent\textbf{Importance of Gated Pooling (GP) Layer.}
We replace GP layer by average pooling  (w/o GP). Tab.~\ref{tab:abla} shows the effectiveness of GP in filtering out noisy information. Instead of treating each element equally, GP help selectively suppress and highlight salient elements within each modality.

\noindent\textbf{Importance of Episode Training.}
We also conduct experiments by removing both CA and GP (w/o GP and CA). In this setting, we concatenate  unimodal representation of images and concepts and use $2$-layer MLP to calculate the relevance score. Although simple, it still achieves satisfactory results, showing episode training is vital for our EpiCA model.

\begin{table}[ht]
\centering
\small
\setlength\tabcolsep{2pt}
\begin{tabular}{ccc}
\toprule
EpiCA variants &  MIT-States(\%) & UT-Zappos(\%)\\ 
\midrule
Full EpiCA & 15.79 & 52.56 \\
- w/o cross attention (CA) & 12.05 & 42.77 \\
- w/o gated pooling (GP) & 13.46 & 50.47 \\
- w/o GP and CA & 14.13 & 48.76\\
\bottomrule
\end{tabular}
\caption{Ablation study of EpiCA components. The episode training and cross-attention encoder are import to our model. Adding gated pooling layer further boosts the accuracy.}
\label{tab:abla}
\end{table}

\begin{figure}
    \centering
    \includegraphics[width=0.4\textwidth]{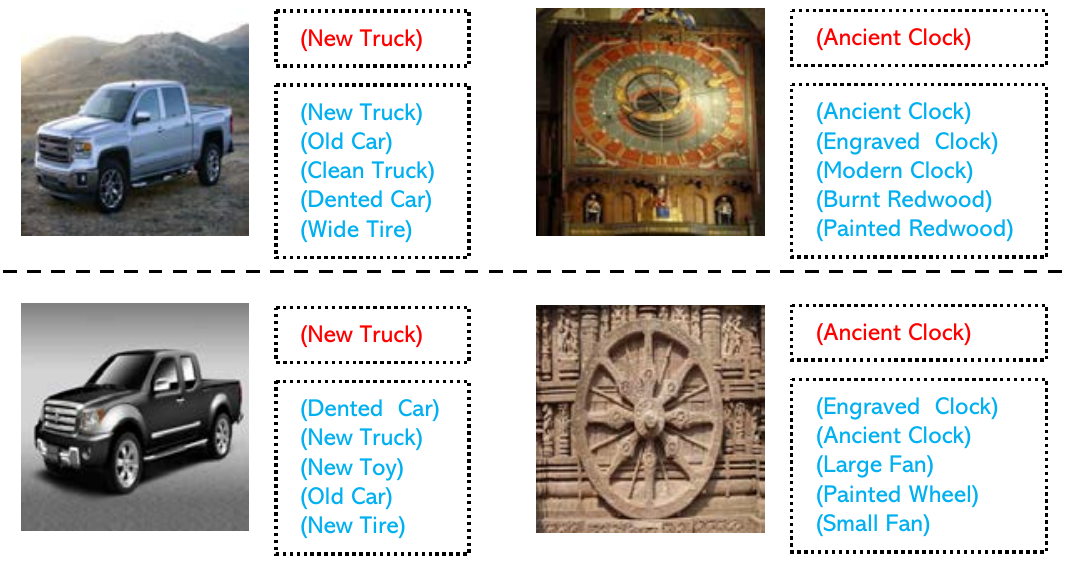}
    \caption{Predicting examples of EpiCA from MIT-States dataset. 
    %Correct predictions are in top row and incorrect predictions are in bottom row. 
    True label and predicted labels are in red and blue text respectively.}
    \label{fig:err_1}
\end{figure}

\subsection{Qualitative Analysis.}
Fig.~\ref{fig:err_1} shows some examples and their  predicted labels by EpiCA. Although it gives the correct predictions for the two examples in the first row, EpiCA still struggles in distinguishing the similar, even opposite attributes, like \textit{New} and \textit{Old}. 
For example, the second highest prediction for the image with true label \textit{new truck} is \textit{old car}. The predicted object is reasonable, but the predicted attribute is opposite. Meanwhile, for the incorrect predictions, the predicted labels are meaningful and remain relevant to the image. For example, \textit{Engraved Clock} may be a better label than \textit{Ancient Clock} for the bottom image. These examples show that EpiCA learns the relevance between images and concepts. But the evaluation of the models is hard and in some cases additional information and bias is needed to predict the exact labels occurring in the dataset. 

\section{Conclusion}
In this paper, we propose EpiCA which combines episode-based training and cross-attention mechanism to exploit the alignment between concepts and images to address ZSCL problems. 
It has led to competitive performance on two benchmark datasets.  In generalized ZSCL setting, EpiCA achieves over $2\times$ performance gain compared to the SOTA on several evaluation metrics. However, ZSCL remains a challenging problem. Future work that explores cognitively motivated learning models and incorporates  information about relations between objects as well as attributes will be interesting directions to pursue.

%The framework is simple with a straightforward architecture and learning objective. 

%\jyc{(do you want to emphasize the improvement in the GZCL setting?I see it almost double the performance of SymNet)}


\bibliographystyle{acl_natbib}
\bibliography{anthology,acl2021}

\begin{thebibliography}{29}
\expandafter\ifx\csname natexlab\endcsname\relax\def\natexlab#1{#1}\fi

\bibitem[{Chen and Grauman(2014)}]{chen2014inferring}
Chao-Yeh Chen and Kristen Grauman. 2014.
\newblock Inferring analogous attributes.
\newblock \emph{Proceedings of the IEEE Conference on Computer Vision and
  Pattern Recognition}, pages 200--207.

\bibitem[{Chen et~al.(2020{\natexlab{a}})Chen, Ding, Liu, Lin, Liu, and
  Han}]{relu_attn}
Hui Chen, Guiguang Ding, Xudong Liu, Zijia Lin, Ji~Liu, and Jungong Han.
  2020{\natexlab{a}}.
\newblock Imram: Iterative matching with recurrent attention memory for
  cross-modal image-text retrieval.
\newblock In \emph{Proceedings of the IEEE/CVF Conference on Computer Vision
  and Pattern Recognition}, pages 12655--12663.

\bibitem[{Chen et~al.(2020{\natexlab{b}})Chen, Gong, and
  Bazzani}]{chen2020image}
Yanbei Chen, Shaogang Gong, and Loris Bazzani. 2020{\natexlab{b}}.
\newblock Image search with text feedback by visiolinguistic attention
  learning.
\newblock pages 3001--3011.

\bibitem[{Chorowski et~al.(2015)Chorowski, Bahdanau, Serdyuk, Cho, and
  Bengio}]{chorowski2015attention}
Jan~K Chorowski, Dzmitry Bahdanau, Dmitriy Serdyuk, Kyunghyun Cho, and Yoshua
  Bengio. 2015.
\newblock Attention-based models for speech recognition.
\newblock \emph{Advances in neural information processing systems},
  28:577--585.

\bibitem[{Dhillon et~al.(2019)Dhillon, Chaudhari, Ravichandran, and
  Soatto}]{dhillon2019baseline}
Guneet~S Dhillon, Pratik Chaudhari, Avinash Ravichandran, and Stefano Soatto.
  2019.
\newblock A baseline for few-shot image classification.
\newblock \emph{arXiv preprint arXiv:1909.02729}.

\bibitem[{He et~al.(2016)He, Zhang, Ren, and Sun}]{he2016deep}
Kaiming He, Xiangyu Zhang, Shaoqing Ren, and Jian Sun. 2016.
\newblock Deep residual learning for image recognition.
\newblock \emph{Proceedings of the IEEE conference on computer vision and
  pattern recognition}, pages 770--778.

\bibitem[{Hochreiter and Schmidhuber(1997)}]{hochreiter1997long}
Sepp Hochreiter and J{\"u}rgen Schmidhuber. 1997.
\newblock Long short-term memory.
\newblock \emph{Neural computation}, 9(8):1735--1780.

\bibitem[{Isola et~al.(2015)Isola, Lim, and Adelson}]{isola2015discovering}
Phillip Isola, Joseph~J Lim, and Edward~H Adelson. 2015.
\newblock Discovering states and transformations in image collections.

\bibitem[{Lake and Baroni(2018)}]{lake2018generalization}
Brenden Lake and Marco Baroni. 2018.
\newblock Generalization without systematicity: On the compositional skills of
  sequence-to-sequence recurrent networks.
\newblock \emph{International Conference on Machine Learning}, pages
  2873--2882.

\bibitem[{Lake et~al.(2015)Lake, Salakhutdinov, and Tenenbaum}]{lake2015human}
Brenden~M Lake, Ruslan Salakhutdinov, and Joshua~B Tenenbaum. 2015.
\newblock Human-level concept learning through probabilistic program induction.
\newblock \emph{Science}, 350(6266):1332--1338.

\bibitem[{Lee et~al.(2018)Lee, Chen, Hua, Hu, and He}]{lee2018stacked}
Kuang-Huei Lee, Xi~Chen, Gang Hua, Houdong Hu, and Xiaodong He. 2018.
\newblock Stacked cross attention for image-text matching.
\newblock pages 201--216.

\bibitem[{Li et~al.(2019{\natexlab{a}})Li, Min, and Fu}]{Li_2019_ICCV}
Kai Li, Martin~Renqiang Min, and Yun Fu. 2019{\natexlab{a}}.
\newblock Rethinking zero-shot learning: A conditional visual classification
  perspective.
\newblock \emph{The IEEE International Conference on Computer Vision (ICCV)}.

\bibitem[{Li et~al.(2019{\natexlab{b}})Li, Sun, Liu, Zhou, Zheng, Chua, and
  Schiele}]{li2019learning}
Xinzhe Li, Qianru Sun, Yaoyao Liu, Qin Zhou, Shibao Zheng, Tat-Seng Chua, and
  Bernt Schiele. 2019{\natexlab{b}}.
\newblock Learning to self-train for semi-supervised few-shot classification.
\newblock \emph{Advances in Neural Information Processing Systems}, pages
  10276--10286.

\bibitem[{Li et~al.(2020)Li, Xu, Mao, and Lu}]{li2020symmetry}
Yong-Lu Li, Yue Xu, Xiaohan Mao, and Cewu Lu. 2020.
\newblock Symmetry and group in attribute-object compositions.
\newblock \emph{Proceedings of the IEEE/CVF Conference on Computer Vision and
  Pattern Recognition}, pages 11316--11325.

\bibitem[{Lu et~al.(2019)Lu, Batra, Parikh, and Lee}]{lu2019vilbert}
Jiasen Lu, Dhruv Batra, Devi Parikh, and Stefan Lee. 2019.
\newblock Vilbert: Pretraining task-agnostic visiolinguistic representations
  for vision-and-language tasks.
\newblock pages 13--23.

\bibitem[{Misra et~al.(2017)Misra, Gupta, and Hebert}]{misra2017red}
Ishan Misra, Abhinav Gupta, and Martial Hebert. 2017.
\newblock From red wine to red tomato: Composition with context.
\newblock \emph{Proceedings of the IEEE Conference on Computer Vision and
  Pattern Recognition}, pages 1792--1801.

\bibitem[{Nagarajan and Grauman(2018)}]{nagarajan2018attrop}
Tushar Nagarajan and Kristen Grauman. 2018.
\newblock Attributes as operators.
\newblock \emph{ECCV}.

\bibitem[{Nan et~al.(2019)Nan, Liu, Zheng, and Zhu}]{nan2019recognizing}
Zhixiong Nan, Yang Liu, Nanning Zheng, and Song-Chun Zhu. 2019.
\newblock Recognizing unseen attribute-object pair with generative model.
\newblock \emph{Proceedings of the AAAI Conference on Artificial Intelligence},
  33:8811--8818.

\bibitem[{Pan and Yang(2009)}]{pan2009survey}
Sinno~Jialin Pan and Qiang Yang. 2009.
\newblock A survey on transfer learning.
\newblock \emph{IEEE Transactions on knowledge and data engineering},
  22(10):1345--1359.

\bibitem[{Pennington et~al.(2014)Pennington, Socher, and
  Manning}]{pennington2014glove}
Jeffrey Pennington, Richard Socher, and Christopher Manning. 2014.
\newblock Glove: Global vectors for word representation.
\newblock \emph{Proceedings of the 2014 conference on empirical methods in
  natural language processing (EMNLP)}, pages 1532--1543.

\bibitem[{Purushwalkam et~al.(2019)Purushwalkam, Nickel, Gupta, and
  Ranzato}]{purushwalkam2019taskdriven}
Senthil Purushwalkam, Maximilian Nickel, Abhinav Gupta, and Marc'Aurelio
  Ranzato. 2019.
\newblock Task-driven modular networks for zero-shot compositional learning.
\newblock \emph{arXiv preprint arXiv:1905.05908}.

\bibitem[{Snell et~al.(2017)Snell, Swersky, and Zemel}]{snell2017prototypical}
Jake Snell, Kevin Swersky, and Richard Zemel. 2017.
\newblock Prototypical networks for few-shot learning.
\newblock \emph{Advances in neural information processing systems}, pages
  4077--4087.

\bibitem[{Sung et~al.(2018)Sung, Yang, Zhang, Xiang, Torr, and
  Hospedales}]{sung2018learning}
Flood Sung, Yongxin Yang, Li~Zhang, Tao Xiang, Philip~HS Torr, and Timothy~M
  Hospedales. 2018.
\newblock Learning to compare: Relation network for few-shot learning.
\newblock \emph{Proceedings of the IEEE Conference on Computer Vision and
  Pattern Recognition}, pages 1199--1208.

\bibitem[{Vaswani et~al.(2017)Vaswani, Shazeer, Parmar, Uszkoreit, Jones,
  Gomez, Kaiser, and Polosukhin}]{vaswani2017attention}
Ashish Vaswani, Noam Shazeer, Niki Parmar, Jakob Uszkoreit, Llion Jones,
  Aidan~N Gomez, {\L}ukasz Kaiser, and Illia Polosukhin. 2017.
\newblock Attention is all you need.
\newblock \emph{Advances in neural information processing systems}, pages
  5998--6008.

\bibitem[{Vinyals et~al.(2016)Vinyals, Blundell, Lillicrap, Wierstra
  et~al.}]{vinyals2016matching}
Oriol Vinyals, Charles Blundell, Timothy Lillicrap, Daan Wierstra, et~al. 2016.
\newblock Matching networks for one shot learning.
\newblock \emph{Advances in neural information processing systems}, pages
  3630--3638.

\bibitem[{Wang et~al.(2019)Wang, Yu, Wang, Darrell, and
  Gonzalez}]{wang2019tafe}
Xin Wang, Fisher Yu, Ruth Wang, Trevor Darrell, and Joseph~E Gonzalez. 2019.
\newblock Tafe-net: Task-aware feature embeddings for low shot learning.
\newblock \emph{Proceedings of the IEEE Conference on Computer Vision and
  Pattern Recognition}, pages 1831--1840.

\bibitem[{Wei et~al.(2019)Wei, Yang, Wang, Deng, and Liu}]{wei2019adversarial}
Kun Wei, Muli Yang, Hao Wang, Cheng Deng, and Xianglong Liu. 2019.
\newblock Adversarial fine-grained composition learning for unseen
  attribute-object recognition.
\newblock pages 3741--3749.

\bibitem[{Yu and Grauman(2017)}]{yu2017semantic}
Aron Yu and Kristen Grauman. 2017.
\newblock Semantic jitter: Dense supervision for visual comparisons via
  synthetic images.
\newblock \emph{Proceedings of the IEEE International Conference on Computer
  Vision}, pages 5570--5579.

\bibitem[{Zhang and Sabuncu(2018)}]{zhang2018generalized}
Zhilu Zhang and Mert Sabuncu. 2018.
\newblock Generalized cross entropy loss for training deep neural networks with
  noisy labels.
\newblock \emph{Advances in neural information processing systems}, pages
  8778--8788.

\end{thebibliography}

%\appendix

\end{document}